\documentclass[letterpaper,10pt,conference]{ieeeconf}

\IEEEoverridecommandlockouts

\usepackage[T1]{fontenc}
\usepackage[utf8]{inputenc}
\usepackage{array}
\usepackage{makecell}
\usepackage{graphicx}
\usepackage{amsmath,amssymb}
\usepackage{booktabs}
\usepackage{tikz}
\usepackage{rotating}
\usepackage{subcaption}
\usepackage[hyphens]{url}
\usepackage{cite}
\usepackage{titlesec}
\titlespacing*{\section}{0pt}{*1.5}{*0.8}
\titlespacing*{\subsection}{0pt}{*1.2}{*0.6}
\usepackage{tabularx}
\usepackage{multirow}
\usepackage{xcolor}
\usepackage{colortbl}
\usepackage{pifont}
\usepackage{threeparttable}
\usepackage{hyperref}
\hypersetup{
    colorlinks=false,
    pdfborder={0 0 1},
    linkbordercolor={0 0.55 0},
    citebordercolor={0 0.55 0},
    urlbordercolor={0 0.55 0}
}
\usepackage{indentfirst}
\usepackage{soul}

\usepackage{aprl_misc}
\usepackage{aprl_acronyms}

\usetikzlibrary{shapes.geometric, arrows.meta, positioning, calc, fit, backgrounds}

\newcommand{\yes}{\textcolor{green!60!black}{\ding{51}}}
\newcommand{\no}{\textcolor{red!70!black}{\ding{55}}}
\newcommand{\pmark}{$\circ$}

\title{\LARGE \bf
Simulation for Planetary Robotic Perception and Autonomy: \\ A Concise Survey of Recent Capabilities and Gaps
}

\author{
Hoyun Kim$^{1*}$ and Giseop Kim$^{1\dagger}$%
\thanks{$^{1}$H. Kim and G. Kim are with the Department of Robotics and
Mechatronics Engineering, DGIST, Daegu, Republic of Korea
{\ttfamily\small \{hoyunkim, gsk\}@dgist.ac.kr}.}%
\thanks{$^{*}$First author: Hoyun Kim. $^{\dagger}$Corresponding author: Giseop Kim.}%
\thanks{This work was supported by the National Research Foundation of Korea (NRF) grants funded by the Korea government (MSIT) (No. RS-2026-25492530 and No. RS-2026-25517444), and by the InnoCORE program of the Ministry of Science and ICT (No. 26-InnoCORE-01).}%
}

\begin{document}

\maketitle
\thispagestyle{empty}
\pagestyle{empty}

\begin{abstract}
Planetary robotics is an important enabler of scientific exploration in
environments where direct human-in-the-loop operation is costly, hazardous, or infeasible.
However, developing and validating planetary robotic systems remains
difficult because representative field testing is expensive, limited, and often
unrepeatable under mission-relevant conditions.
In this setting, simulation serves as a central tool for perception
and autonomy research, synthetic data generation, system integration, and
pre-deployment evaluation. Despite its importance, the literature on planetary
robotics simulation remains dispersed across different simulation engines,
implementations, and application settings.
This paper surveys simulation works for planetary robotic perception and
autonomy across four practical axes: Openness and Availability, Scenario and
Platform Coverage, Sensor and Perception Support, and Environmental and
Operational Realism.
The surveyed simulation works report visual or physical fidelity and support
perception-oriented workflows. They also indicate uneven public availability,
rover-centered coverage, partial support for specialized sensing modalities, and
uneven reporting of operational constraints such as onboard computation,
energy, and communication restrictions.
\end{abstract}

\section{Introduction}

Space robotics covers robotic systems designed to operate beyond Earth.
Prior reviews distinguish orbital and planetary applications as different
autonomy contexts \cite{YGao2017}: orbital robots typically operate in
microgravity around spacecraft, whereas planetary robots must traverse natural
surfaces under uncertain terrain, variable illumination, limited communication,
and long-duration operational constraints \cite{YGao2017}, \cite{FIngrand2007}.
This paper focuses on the latter class of systems, namely robots intended for
planetary surface or near-surface exploration. That class is also broadening
beyond wheeled rovers to include legged, aerial, and heterogeneous robotic
systems \cite{PArm2023}, \cite{RDomingues2025}.

Simulation is central in this setting because representative validation is
expensive, limited, and difficult to repeat. Planetary autonomy also couples
perception, localization, motion generation, and decision-making
\cite{FIngrand2007}; recent work on perception-aware planning, global
localization, and multi-robot coordination further illustrates this coupling
\cite{Strader2020,Geromichalos2020,Censible2024,RDomingues2025}.
As a result, simulation works increasingly support more than visualization:
they are used for synthetic data generation, perception-oriented evaluation,
and software-in-the-loop experimentation \cite{Mueller2021}, \cite{Richard_2024}.


\definecolor{colCardTerra}{HTML}{C1664B}   
\definecolor{colCardAmber}{HTML}{D9A441}   
\definecolor{colCardIndigo}{HTML}{5C6BAB}  
\definecolor{colCardTeal}{HTML}{3E8E8B}    

\newcommand{\bandcard}[5]{%
    \fill[white, rounded corners=6pt] (#1,#2) rectangle (#3,#4);
    \fill[#5!12, rounded corners=6pt] (#1,#2) rectangle (#3,#2-0.75);
    \fill[#5!12] (#1,#2-0.38) rectangle (#3,#2-0.75);
    \draw[cardS, draw=#5!55] (#1,#2) rectangle (#3,#4);
}
\newcommand{\chip}[6]{%
    \fill[#5!7, rounded corners=3pt] (#1,#2) rectangle (#1+#3,#2-#4);
    \draw[chipS, draw=#5!45] (#1,#2) rectangle (#1+#3,#2-#4);
    \node[chiptxt] at (#1+#3/2, #2-#4/2) {#6};
}
\newcommand{\bandchev}[3]{%
    \draw[line width=1.8pt, #2, line cap=round, line join=round]
        (5.35,#1) -- (5.60,#1-0.25) -- (5.85,#1);
    \node[chevlab] at (6.05, #1-0.12) {#3};
}

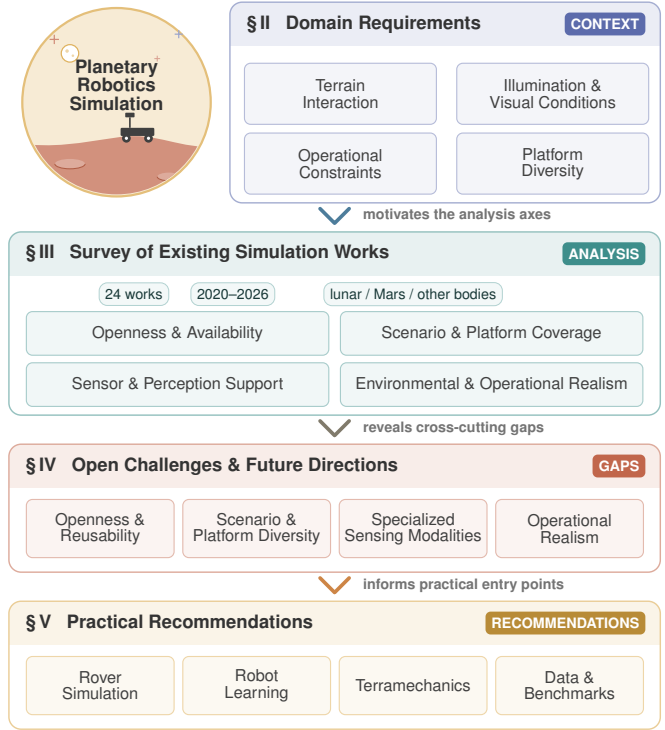
\begin{figure}[!t]
\centering
\resizebox{\columnwidth}{!}{%
\begin{tikzpicture}[
    x=1cm, y=1cm,
    every node/.style={inner sep=1pt},
    cardS/.style={rounded corners=6pt, line width=0.7pt},
    chipS/.style={rounded corners=3pt, line width=0.55pt},
    bandtitle/.style={font=\sffamily\small\bfseries, text=black!80,
        anchor=west, align=left},
    tagchip/.style={rounded corners=2.5pt, font=\sffamily\scriptsize\bfseries,
        text=white, inner sep=3pt, anchor=east},
    chiptxt/.style={font=\sffamily\footnotesize, text=black!75, align=center},
    pill/.style={rounded corners=4pt, draw=colCardTeal!40, fill=colCardTeal!6,
        font=\sffamily\scriptsize, text=colCardTeal!40!black, inner sep=3pt},
    chevlab/.style={font=\sffamily\scriptsize\bfseries, text=black!55,
        anchor=west},
]

\draw[fill=colCardAmber!22, draw=colCardAmber!70, line width=1.1pt]
    (1.85,-1.72) circle (1.62);
\begin{scope}
    \clip (1.85,-1.72) circle (1.585);
    \fill[colCardTerra!72] (0.10,-2.42)
        to[out=-8, in=188] (1.90,-2.32)
        to[out=8, in=185] (3.60,-2.48)
        -- (3.60,-3.45) -- (0.10,-3.45) -- cycle;
    \fill[colCardTerra!55] (1.05,-2.78) ellipse (0.26 and 0.085);
    \draw[colCardTerra!90!black, line width=0.5pt]
        (0.79,-2.78) arc[start angle=180, end angle=360, x radius=0.26, y radius=0.085];
    \fill[colCardTerra!55] (2.75,-3.00) ellipse (0.20 and 0.07);
    \draw[colCardTerra!90!black, line width=0.5pt]
        (2.55,-3.00) arc[start angle=180, end angle=360, x radius=0.20, y radius=0.07];
\end{scope}
\fill[black!75, rounded corners=0.6pt] (1.92,-2.16) rectangle (2.50,-2.30);
\fill[black!75] (2.02,-2.34) circle (0.075);
\fill[black!75] (2.40,-2.34) circle (0.075);
\fill[white] (2.02,-2.34) circle (0.028);
\fill[white] (2.40,-2.34) circle (0.028);
\draw[black!75, line width=0.8pt] (2.08,-2.16) -- (2.08,-1.98);
\fill[black!75, rounded corners=0.4pt] (2.00,-1.98) rectangle (2.16,-1.90);
\draw[colCardTerra!75, line width=0.6pt] (0.78,-0.72) -- (0.78,-0.56)
    (0.70,-0.64) -- (0.86,-0.64);
\draw[colCardIndigo!60, line width=0.6pt] (2.92,-0.62) -- (2.92,-0.48)
    (2.85,-0.55) -- (2.99,-0.55);
\draw[colCardTerra!60, line width=0.5pt] (2.55,-1.02) -- (2.55,-0.92)
    (2.50,-0.97) -- (2.60,-0.97);
\fill[white] (1.05,-0.88) circle (0.15);
\draw[colCardAmber!85, line width=0.6pt] (1.05,-0.88) circle (0.15);
\fill[colCardAmber!45] (1.00,-0.84) circle (0.030);
\fill[colCardAmber!45] (1.10,-0.92) circle (0.022);
\node[align=center, font=\sffamily\small\bfseries, text=black!80]
    at (1.85,-1.42)
    {Planetary\\[-1pt]Robotics\\[-1pt]Simulation};

\bandcard{3.8}{0}{11.2}{-3.45}{colCardIndigo}
\node[bandtitle] at (4.05,-0.375)
    {\S\,\ref*{sec:prelim}\;\; Domain Requirements};
\node[tagchip, fill=colCardIndigo] at (10.95,-0.375) {CONTEXT};
\chip{4.05}{-1.05}{3.3}{1.05}{colCardIndigo}{Terrain\\[-1pt]Interaction}
\chip{7.70}{-1.05}{3.3}{1.05}{colCardIndigo}{Illumination \&\\[-1pt]Visual Conditions}
\chip{4.05}{-2.25}{3.3}{1.05}{colCardIndigo}{Operational\\[-1pt]Constraints}
\chip{7.70}{-2.25}{3.3}{1.05}{colCardIndigo}{Platform\\[-1pt]Diversity}

\bandchev{-3.55}{colCardIndigo!50!colCardTeal}{motivates the analysis axes}

\bandcard{0}{-3.95}{11.2}{-7.1}{colCardTeal}
\node[bandtitle] at (0.25,-4.325)
    {\S\,\ref*{sec:survey}\;\; Survey of Existing Simulation Works};
\node[tagchip, fill=colCardTeal] at (10.95,-4.325) {ANALYSIS};
\node[pill] at (2.15,-5.02) {24 works};
\node[pill] at (3.85,-5.02) {2020--2026};
\node[pill] at (6.95,-5.02) {lunar~/~Mars~/~other bodies};
\chip{0.30}{-5.32}{5.2}{0.75}{colCardTeal}{Openness \& Availability}
\chip{5.70}{-5.32}{5.2}{0.75}{colCardTeal}{Scenario \& Platform Coverage}
\chip{0.30}{-6.20}{5.2}{0.75}{colCardTeal}{Sensor \& Perception Support}
\chip{5.70}{-6.20}{5.2}{0.75}{colCardTeal}{Environmental \& Operational Realism}

\bandchev{-7.2}{colCardTeal!50!colCardTerra}{reveals cross-cutting gaps}

\bandcard{0}{-7.6}{11.2}{-9.8}{colCardTerra}
\node[bandtitle] at (0.25,-7.975)
    {\S\,\ref*{sec:challenges}\;\; Open Challenges \& Future Directions};
\node[tagchip, fill=colCardTerra] at (10.95,-7.975) {GAPS};
\chip{0.30}{-8.55}{2.54}{1.0}{colCardTerra}{Openness \&\\[-1pt]Reusability}
\chip{2.99}{-8.55}{2.54}{1.0}{colCardTerra}{Scenario \&\\[-1pt]Platform Diversity}
\chip{5.68}{-8.55}{2.54}{1.0}{colCardTerra}{Specialized\\[-1pt]Sensing Modalities}
\chip{8.37}{-8.55}{2.54}{1.0}{colCardTerra}{Operational\\[-1pt]Realism}

\bandchev{-9.9}{colCardTerra!50!colCardAmber}{informs practical entry points}

\bandcard{0}{-10.3}{11.2}{-12.5}{colCardAmber}
\node[bandtitle] at (0.25,-10.675)
    {\S\,\ref*{sec:ready}\;\; Practical Recommendations};
\node[tagchip, fill=colCardAmber!85!black] at (10.95,-10.675) {RECOMMENDATIONS};
\chip{0.30}{-11.25}{2.54}{1.0}{colCardAmber}{Rover\\[-1pt]Simulation}
\chip{2.99}{-11.25}{2.54}{1.0}{colCardAmber}{Robot\\[-1pt]Learning}
\chip{5.68}{-11.25}{2.54}{1.0}{colCardAmber}{Terramechanics}
\chip{8.37}{-11.25}{2.54}{1.0}{colCardAmber}{Data \&\\[-1pt]Benchmarks}

\end{tikzpicture}%
}%
\caption{Structure of this survey. The paper first derives planetary
simulation requirements, then surveys twenty-four simulation works along four
analysis axes, summarizes the resulting gaps as open challenges, and provides
practical starting points for new users.}
\label{fig:overview}
\vspace{-8pt}
\end{figure}


\definecolor{colMars}{HTML}{C9763B}    
\definecolor{colLunar}{HTML}{75787E}   
\definecolor{colPlanet}{HTML}{2EA89A}  
\definecolor{colEvent}{HTML}{A93226}   

\newcommand{\yrblk}[5]{%
    \fill[#4, rounded corners=2.5pt] (#1+0.03, #3-0.25) rectangle (#2-0.03, #3+0.25);
    \node[font=\sffamily\footnotesize\bfseries, text=white] at ({(#1+#2)/2}, #3) {#5};
}
\newcommand{\axbreak}[2]{%
    \draw[#2!75, line width=1.0pt]
        (3.95, #1-0.42) .. controls (3.75, #1-0.14) and (4.15, #1+0.14) .. (3.95, #1+0.42);
    \draw[#2!75, line width=1.0pt]
        (4.26, #1-0.42) .. controls (4.06, #1-0.14) and (4.46, #1+0.14) .. (4.26, #1+0.42);
}
\newcommand{\stem}[4]{\draw[#4!50, line width=0.7pt] (#1,#2) -- (#1,#3);}

\begin{figure*}[!tp]
\centering
\resizebox{\textwidth}{!}{%
\begin{tikzpicture}[
    x=1cm, y=1cm,
    every node/.style={inner sep=1pt},
    lan/.style={font=\sffamily\small\bfseries, rotate=90},
    lab/.style={font=\sffamily\footnotesize, text=black!75, align=center},
    labE/.style={font=\sffamily\scriptsize, text=colEvent, anchor=west},
    labT/.style={font=\sffamily\scriptsize, text=black!55, align=center},
    leg/.style={font=\sffamily\scriptsize, text=black!65, anchor=west,
        text height=5.5pt, text depth=1.5pt},
    legtitle/.style={leg, font=\sffamily\scriptsize\bfseries, text=black!70},
    cirf/.style={circle, fill=#1, minimum size=5.5pt, inner sep=0pt},
    cirh/.style={circle, draw=#1, fill=white, line width=0.9pt, minimum size=5.5pt, inner sep=0pt},
    diaf/.style={diamond, fill=#1, minimum size=7pt, inner sep=0pt},
    diah/.style={diamond, draw=#1, fill=white, line width=0.9pt, minimum size=7pt, inner sep=0pt},
    trif/.style={regular polygon, regular polygon sides=3, fill=#1, minimum size=7pt, inner sep=0pt},
    trih/.style={regular polygon, regular polygon sides=3, draw=#1, fill=white, line width=0.9pt, minimum size=7pt, inner sep=0pt},
    pentf/.style={regular polygon, regular polygon sides=5, fill=#1, minimum size=7pt, inner sep=0pt},
    starf/.style={star, star points=5, fill=#1, minimum size=8pt, inner sep=0pt},
    anc/.style={rectangle, rounded corners=1pt, draw=#1, fill=#1!25, minimum size=6pt, inner sep=0pt, line width=0.9pt},
    evt/.style={rectangle, fill=colEvent, minimum size=5pt, inner sep=0pt},
    eng/.style={rectangle, draw=black!55, fill=white, minimum size=5pt, inner sep=0pt, line width=0.6pt},
]


\fill[colLunar!12, rounded corners=3pt] (0, 6.7) rectangle (0.5, 9.5);
\node[lan, text=colLunar] at (0.25, 8.1) {Lunar};
\fill[colPlanet!12, rounded corners=3pt] (0, 3.6) rectangle (0.5, 6.6);
\node[lan, text=colPlanet!80!black] at (0.25, 5.1) {Cross-domain};
\fill[colMars!12, rounded corners=3pt] (0, 0.8) rectangle (0.5, 3.4);
\node[lan, text=colMars] at (0.25, 2.1) {Mars};

\yrblk{0.7}{3.8}{8.1}{colLunar!85}{1960s}
\axbreak{8.1}{colLunar}
\yrblk{4.45}{7.05}{8.1}{colLunar!55}{2020}
\yrblk{7.05}{9.75}{8.1}{colLunar!85}{2021}
\yrblk{9.75}{11.45}{8.1}{colLunar!55}{2022}
\yrblk{11.45}{14.85}{8.1}{colLunar!85}{2023}
\yrblk{14.85}{18.45}{8.1}{colLunar!55}{2024}
\yrblk{18.45}{21.95}{8.1}{colLunar!85}{2025}
\yrblk{21.95}{24.65}{8.1}{colLunar!55}{2026}
\draw[colLunar!85, line width=1.2pt, -{Stealth[scale=1.0]}] (24.68, 8.1) -- (25.05, 8.1);
\yrblk{0.7}{3.8}{5.1}{colPlanet!85}{2002--05}
\axbreak{5.1}{colPlanet}
\yrblk{4.45}{7.05}{5.1}{colPlanet!55}{2020}
\yrblk{7.05}{9.75}{5.1}{colPlanet!85}{2021}
\yrblk{9.75}{11.45}{5.1}{colPlanet!55}{2022}
\yrblk{11.45}{14.85}{5.1}{colPlanet!85}{2023}
\yrblk{14.85}{18.45}{5.1}{colPlanet!55}{2024}
\yrblk{18.45}{21.95}{5.1}{colPlanet!85}{2025}
\yrblk{21.95}{24.65}{5.1}{colPlanet!55}{2026}
\draw[colPlanet!85, line width=1.2pt, -{Stealth[scale=1.0]}] (24.68, 5.1) -- (25.05, 5.1);
\yrblk{0.7}{3.8}{2.1}{colMars!85}{2003}
\axbreak{2.1}{colMars}
\yrblk{4.45}{7.05}{2.1}{colMars!55}{2020}
\yrblk{7.05}{9.75}{2.1}{colMars!85}{2021}
\yrblk{9.75}{11.45}{2.1}{colMars!55}{2022}
\yrblk{11.45}{14.85}{2.1}{colMars!85}{2023}
\yrblk{14.85}{18.45}{2.1}{colMars!55}{2024}
\yrblk{18.45}{21.95}{2.1}{colMars!85}{2025}
\yrblk{21.95}{24.65}{2.1}{colMars!55}{2026}
\draw[colMars!85, line width=1.2pt, -{Stealth[scale=1.0]}] (24.68, 2.1) -- (25.05, 2.1);

\stem{2.25}{8.35}{8.65}{colLunar}
\node[anc=colLunar] at (2.25, 8.65) {};
\node[lab] at (2.25, 9.12) {Apollo LM mission\\simulators};
\stem{10.5}{8.35}{9.35}{colLunar} \node[pentf=colLunar] at (10.5, 9.35) {};
\node[lab] at (10.5, 9.65) {GraspPlanetary~\cite{GraspPlanetary2022}};
\stem{12.3}{8.35}{8.65}{colLunar} \node[diah=colLunar] at (12.3, 8.65) {};
\node[lab] at (12.3, 8.95) {SEELO~\cite{Cloud2023}};
\stem{15.7}{8.35}{8.65}{colLunar} \node[cirf=colLunar] at (15.7, 8.65) {};
\node[lab] at (15.7, 8.95) {OmniLRS~\cite{Richard_2024}};
\stem{19.2}{8.35}{8.65}{colLunar} \node[cirh=colLunar] at (19.2, 8.65) {};
\node[lab] at (19.2, 8.95) {Lindmark et al.~\cite{Daniel2025}};
\stem{22.6}{8.35}{8.65}{colLunar} \node[cirf=colLunar] at (22.6, 8.65) {};
\node[lab] at (22.6, 8.95) {POLAR-Sim~\cite{POLARSim2026}};
\stem{13.8}{8.35}{9.35}{colLunar} \node[cirf=colLunar] at (13.8, 9.35) {};
\node[lab] at (13.8, 9.65) {LunarSim~\cite{LunarSim2023}};
\stem{18.9}{8.35}{9.35}{colLunar} \node[diaf=colLunar] at (18.9, 9.35) {};
\node[lab] at (18.9, 9.65) {Batagoda et al.~\cite{Chrono2024}};
\stem{21.0}{8.35}{10.05}{colLunar} \node[diah=colLunar] at (21.0, 10.05) {};
\node[lab] at (21.0, 10.35) {Linde et al.~\cite{Mattias2025}};
\stem{13.2}{7.85}{7.55}{colLunar}
\node[cirh=colLunar] at (13.2, 7.55) {};
\fill[colLunar] (13.2, 7.55) ++(0, 0.075) arc[start angle=90, end angle=270, radius=0.075] -- cycle;
\node[lab] at (13.2, 7.25) {DUST~\cite{DUST2023}};
\stem{16.6}{7.85}{7.55}{colLunar} \node[cirf=colLunar] at (16.6, 7.55) {};
\node[lab] at (16.6, 7.25) {LunarRoverSim~\cite{LunarRoverSim2024}};
\stem{20.1}{7.85}{7.55}{colLunar} \node[cirf=colLunar] at (20.1, 7.55) {};
\node[lab] at (20.1, 7.25) {LuSeg/LESS~\cite{LuSeg2025}};
\stem{23.5}{7.85}{7.55}{colLunar} \node[cirh=colLunar] at (23.5, 7.55) {};
\node[lab] at (23.5, 7.25) {Kurt et al.~\cite{DualSim2026}};

\stem{1.6}{5.35}{5.65}{colPlanet}
\node[eng] at (1.6, 5.65) {};
\node[labT] at (1.6, 5.95) {Gazebo (2002)};
\stem{2.9}{5.35}{6.35}{colPlanet}
\node[eng] at (2.9, 6.35) {};
\node[labT] at (2.9, 6.65) {Unity (2005)};
\stem{7.9}{5.35}{5.65}{colPlanet} \node[starf=colPlanet] at (7.9, 5.65) {};
\node[lab] at (7.9, 5.97) {OAISYS~\cite{Mueller2021}};
\stem{5.6}{5.35}{6.35}{black} \node[eng] at (5.6, 6.35) {};
\node[labT] at (5.6, 6.65) {Isaac Sim (2020)};
\stem{10.9}{5.35}{5.65}{black} \node[eng] at (10.9, 5.65) {};
\node[labT] at (10.9, 5.95) {Unreal Engine 5 (2022)};
\stem{15.6}{5.35}{5.65}{colPlanet} \node[cirf=colPlanet] at (15.6, 5.65) {};
\node[lab] at (15.6, 5.95) {RLRoverLab~\cite{RLRoverLAB2024}};
\stem{17.4}{5.35}{6.35}{colPlanet} \node[cirf=colPlanet] at (17.4, 6.35) {};
\node[lab] at (17.4, 6.65) {GUISS~\cite{GUISS2024}};
\stem{19.8}{5.35}{5.65}{colPlanet} \node[cirf=colPlanet] at (19.8, 5.65) {};
\node[lab] at (19.8, 5.95) {PlanetaryPathBench~\cite{PlanetaryPathBench2025}};
\stem{21.1}{4.85}{4.40}{colPlanet} \node[starf=colPlanet] at (21.1, 4.40) {};
\node[lab] at (21.1, 4.10) {SRB~\cite{SRB2025}};
\stem{4.55}{4.85}{4.48}{colEvent} \node[evt] at (4.55, 4.48) {};
\node[labE] at (4.71, 4.48) {Chang'e-4 (2019)};
\stem{9.95}{4.85}{4.48}{colEvent} \node[evt] at (9.95, 4.48) {};
\node[labE] at (10.11, 4.48) {Artemis I launch (2022)};
\stem{14.95}{4.85}{4.48}{colEvent} \node[evt] at (14.95, 4.48) {};
\node[labE] at (15.11, 4.48) {Chang'e-6 sample return (2024)};
\stem{7.35}{4.85}{3.96}{colEvent} \node[evt] at (7.35, 3.96) {};
\node[labE] at (7.51, 3.96) {Perseverance \& Ingenuity (2021)};

\stem{2.25}{2.35}{2.65}{colMars}
\node[anc=colMars] at (2.25, 2.65) {};
\node[lab] at (2.25, 2.95) {ROAMS (NASA JPL)};
\stem{5.7}{2.35}{2.65}{colMars} \node[cirf=colMars] at (5.7, 2.65) {};
\node[lab] at (5.7, 2.95) {Giubilato et al.~\cite{Giubilato2020}};
\stem{12.5}{2.35}{2.65}{colMars} \node[cirh=colMars] at (12.5, 2.65) {};
\node[lab] at (12.5, 2.95) {MarsSim~\cite{MarsSim2023}};
\stem{15.9}{2.35}{2.65}{colMars} \node[trih=colMars] at (15.9, 2.65) {};
\node[lab] at (15.9, 2.95) {MarsDrone~\cite{MarsDrone2024}};
\stem{22.9}{2.35}{2.65}{colMars} \node[trif=colMars] at (22.9, 2.65) {};
\node[lab] at (22.9, 2.95) {MARTIAN~\cite{MARTIAN2026}};
\stem{6.4}{1.85}{1.55}{colMars} \node[cirh=colMars] at (6.4, 1.55) {};
\node[lab] at (6.4, 1.25) {Toupet et al.~\cite{Toupet2020ENavSim}};
\stem{13.7}{1.85}{1.55}{colMars} \node[cirh=colMars] at (13.7, 1.55) {};
\node[lab] at (13.7, 1.25) {Jiang et al.~\cite{Yunn2023}};
\stem{17.1}{1.85}{1.55}{colMars} \node[cirh=colMars] at (17.1, 1.55) {};
\node[lab] at (17.1, 1.25) {ISMRS~\cite{ISMRS2024}};

\draw[black!45, line width=0.8pt] (0.7, 0.72) -- (0.7, 0.60) -- (3.8, 0.60) -- (3.8, 0.72);
\node[font=\sffamily\scriptsize\bfseries, text=black!55] at (2.25, 0.36) {Mission-era tools};
\draw[black!45, line width=0.8pt] (4.45, 0.72) -- (4.45, 0.60) -- (24.65, 0.60) -- (24.65, 0.72);
\node[font=\sffamily\scriptsize\bfseries, text=black!55] at (14.55, 0.36)
    {Recent simulation works (24 surveyed works)};

\node[legtitle] at (0.7, -0.25) {Platform:};
\node[cirf=black!55] at (2.15, -0.25) {};
\node[leg] at (2.28, -0.25) {rover-relevant};
\node[diaf=black!55] at (4.35, -0.25) {};
\node[leg] at (4.48, -0.25) {construction};
\node[trif=black!55] at (6.40, -0.25) {};
\node[leg] at (6.53, -0.25) {aerial};
\node[starf=black!55] at (7.75, -0.25) {};
\node[leg] at (7.88, -0.25) {multi-platform};
\node[pentf=black!55] at (9.80, -0.25) {};
\node[leg] at (9.93, -0.25) {manipulation};
\node[legtitle] at (11.65, -0.25) {Access:};
\node[cirf=black!55] at (12.80, -0.25) {};
\node[leg] at (12.93, -0.25) {public release};
\node[cirh=black!55] at (14.75, -0.25) {};
\node[leg] at (14.88, -0.25) {not identified};
\node[cirh=black!55] at (16.65, -0.25) {};
\fill[black!55] (16.65, -0.25) ++(0, 0.075) arc[start angle=90, end angle=270, radius=0.075] -- cycle;
\node[leg] at (16.78, -0.25) {on request};
\node[evt] at (18.65, -0.25) {};
\node[leg] at (18.78, -0.25) {space-mission event};
\node[eng] at (21.75, -0.25) {};
\node[leg] at (21.88, -0.25) {engine milestone};

\end{tikzpicture}%
}%
\caption{Chronological roadmap of the surveyed planetary simulation works as
domain lanes (top: lunar; middle: cross-domain or other bodies; bottom:
Mars). Each lane starts from an early mission-era anchor (Apollo LM mission
simulators, the ROAMS rover simulator at NASA JPL, and selected development
and release milestones of general-purpose engines; shown as uncited historical markers), and the
wavy break compresses the intervening decades before the recent time series.
Marker shape encodes the platform or task category, and marker fill encodes
simulation resource availability (filled: public release; outline: no release
identified; half: on request). Further details on the surveyed works are
provided in Table~\ref{tab:sim-profile}. Dark-red squares mark major space-mission events and open gray
squares mark simulation-engine milestones.}
\label{fig:roadmap}
\vspace{-8pt}
\end{figure*}
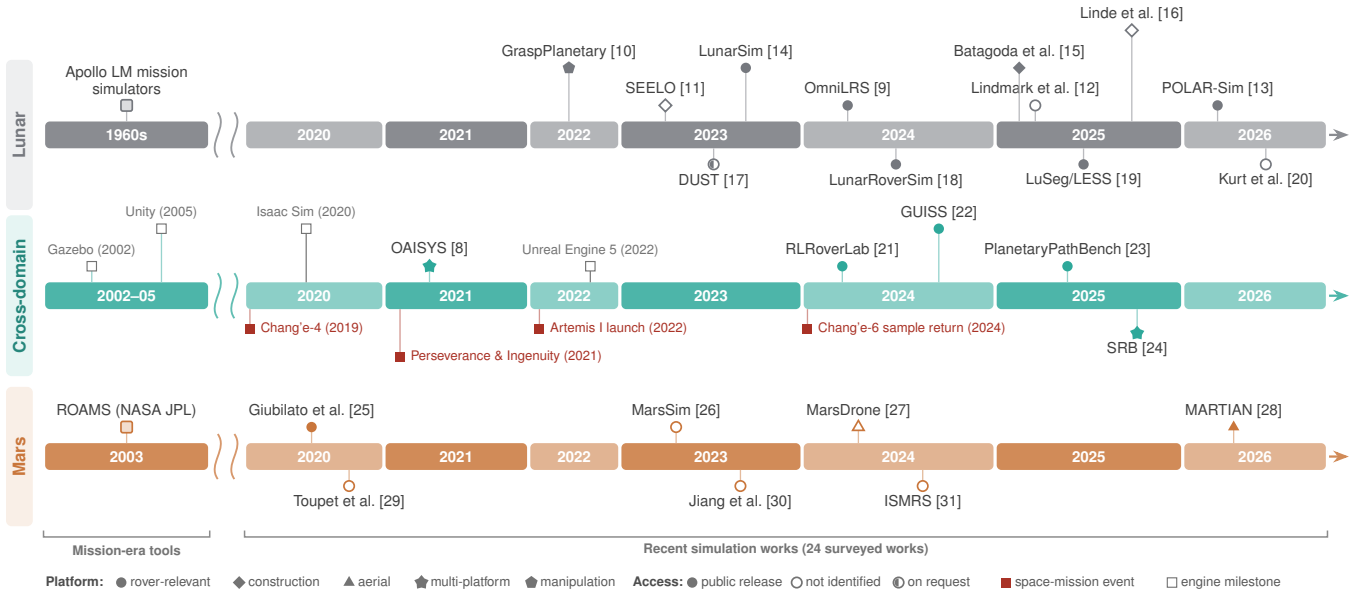

Previous reviews and system studies cover important parts of the broader field, including space
robotics \cite{YGao2017}, decisional autonomy \cite{FIngrand2007}, active
SLAM \cite{Julio2023}, and AI-enabled vision for space missions \cite{Quoos2026}.
Complementing these perspectives, we compare simulation works
in terms of their reported capabilities and resource availability.
Such a comparison is useful because simulator choice depends not only on
the target algorithm or mission, but also on sensing modalities, robotic
platforms, synthetic data needs, environmental conditions, and autonomy-stack
compatibility. Accordingly, we survey simulation works as they are
used in planetary perception and autonomy research. In this paper, simulation
works include simulators, rendering pipelines, synthetic-data generators, and
benchmark-oriented environments when they support planetary perception or
autonomy research.

This paper addresses that need with three contributions:
\begin{enumerate}
    \item We provide a simulation-centered perspective on recent works for planetary robotic perception and autonomy, organized around four practical axes: openness, scenario coverage, sensing support, and operational realism.
    \item We relate these axes to planetary autonomy requirements and summarize the reported capabilities of twenty-four surveyed works across access, platform scope, sensing, task outputs, and environmental or operational assumptions.
    \item We identify recurring gaps in this mapping and discuss future directions toward reusable planetary simulation ecosystems with explicit assumptions.
\end{enumerate}

The remainder of this paper is organized as follows.
Section~\ref{sec:prelim} introduces the mission context and derives the
simulation requirements that motivate the survey axes.
Section~\ref{sec:survey} surveys existing simulation works from four analytical
perspectives: Openness and Availability, Scenario and Platform Coverage, Sensor
and Perception Support, and Environmental and Operational Realism.
Section~\ref{sec:challenges} discusses open challenges and future directions,
Section~\ref{sec:ready} distills practical simulator recommendations for
newcomers, and Section~\ref{sec:conclusion} concludes the paper.
Fig.~\ref{fig:overview} summarizes this structure, and Fig.~\ref{fig:roadmap}
presents the twenty-four surveyed simulation works chronologically by target
domain.

\section{Planetary Robotics Preliminaries}
\label{sec:prelim}


\definecolor{colCardTerra}{HTML}{C1664B}   
\definecolor{colCardAmber}{HTML}{D9A441}   
\definecolor{colCardIndigo}{HTML}{5C6BAB}  
\definecolor{colCardTeal}{HTML}{3E8E8B}    

\newcommand{\factrow}[5]{%
    \fill[#2, rounded corners=1.5pt] (0.15, #1-0.13) rectangle (0.41, #1+0.13);
    \node[factortxt] at (0.62, #1) {#3};
    \draw[cardC, draw=#2!50, fill=white] (5.3, #1-0.48) rectangle (14.1, #1+0.48);
    \node[chartxt] at (5.6, #1) {#4};
    \draw[line width=1.8pt, #2, line cap=round, line join=round]
        (14.32, #1+0.20) -- (14.60, #1) -- (14.32, #1-0.20);
    \draw[cardC, draw=#2!45, fill=#2!10] (14.95, #1-0.48) rectangle (23.7, #1+0.48);
    \node[impltxt, text=#2!45!black] at (15.25, #1) {#5};
}

\begin{figure*}[!tp]
\centering
\resizebox{\textwidth}{!}{%
\begin{tikzpicture}[
    x=1cm, y=1cm,
    every node/.style={inner sep=1pt},
    colhead/.style={font=\sffamily\small\bfseries, text=black!60, anchor=west},
    grouplab/.style={font=\sffamily\scriptsize\bfseries, anchor=west,
        fill=white, inner sep=3pt},
    factortxt/.style={font=\sffamily\small\bfseries, text=black!80,
        anchor=west, align=left, text width=4.45cm},
    chartxt/.style={font=\sffamily\small, text=black!75,
        anchor=west, align=left, text width=8.0cm},
    impltxt/.style={font=\sffamily\small\bfseries,
        anchor=west, align=left, text width=8.0cm},
    cardC/.style={rounded corners=3pt, line width=0.7pt},
]

\node[colhead] at (0.15, 0)  {Domain Factor};
\node[colhead] at (5.6, 0)   {Planetary Characteristic};
\node[colhead] at (15.25, 0) {Simulation Implication (This Survey)};
\draw[black!15, line width=0.6pt] (0.15, -0.40) -- (23.7, -0.40);

\draw[colCardTerra!40, line width=0.8pt] (0.15, -0.95) -- (23.7, -0.95);
\node[grouplab, text=colCardTerra!80!black] at (0.15, -0.95) {PHYSICAL ENVIRONMENT};

\factrow{-1.70}{colCardTerra}
    {Terrain interaction}
    {Slip, sinkage, and wheel--regolith coupling}
    {Physics-based terramechanics}
\factrow{-2.80}{colCardTerra}
    {Illumination and visual conditions}
    {Low texture and extreme illumination}
    {Physically-based rendering and sensor modeling}
\factrow{-3.90}{colCardTerra}
    {Gravity and contact regime}
    {Low-gravity regimes}
    {Configurable gravity and contact dynamics}

\draw[colCardIndigo!40, line width=0.8pt] (0.15, -4.65) -- (23.7, -4.65);
\node[grouplab, text=colCardIndigo!80!black] at (0.15, -4.65) {OPERATIONAL CONSTRAINTS};

\factrow{-5.40}{colCardIndigo}
    {Communication}
    {Communication delay and limited bandwidth}
    {Latency-aware autonomy modeling}
\factrow{-6.50}{colCardIndigo}
    {Onboard resources}
    {Bounded power, compute, and mission duration}
    {Energy- and compute-aware evaluation}

\draw[colCardTeal!40, line width=0.8pt] (0.15, -7.25) -- (23.7, -7.25);
\node[grouplab, text=colCardTeal!80!black] at (0.15, -7.25) {ROBOTIC SYSTEM SCOPE};

\factrow{-8.00}{colCardTeal}
    {Platform diversity}
    {Rovers, legged robots, aerial vehicles, and multi-robot systems}
    {Heterogeneous platform support}
\factrow{-9.10}{colCardTeal}
    {Sensor and autonomy stack}
    {Integrated perception, localization, planning, and resource management}
    {Task-relevant sensing and stack integration}

\end{tikzpicture}%
}%
\caption{Key factors distinguishing planetary from terrestrial autonomy and
their implications for simulation design. Each domain factor (left) is linked
to its reported planetary characteristic (center) and the simulation
capability it implies (right), grouped into physical environment, operational
constraints, and robotic system scope.}
\label{fig:domain-diff}
\vspace{-8pt}
\end{figure*}

\subsection{Planetary Robotics as an Integrated Autonomy Problem}

Planetary robotics concerns robotic operation on planetary surfaces. In this
setting, systems must handle uncertain terrain interaction, including slip,
sinkage, and unknown surface properties, as well as significant communication
delays with Earth and GNSS-denied localization. They must also operate under
extreme environmental constraints, including radiation, wide thermal cycling,
and body-dependent atmospheric conditions, during missions that can span
months to years with communication delays limiting real-time intervention.
Early work already treated perception,
localization, motion
generation, temporal planning, and resource management as components of one
operational system \cite{FIngrand2007}, and later studies on terrain-adaptive
navigation, global localization, and perception-aware planning reinforce this
coupling \cite{Strader2020,Lacroix2002,Ishigami2013,Gonzalez2018,Geromichalos2020,Censible2024}.

These constraints distinguish planetary robotics from terrestrial autonomy in
ways that directly affect simulation design. Fig.~\ref{fig:domain-diff}
summarizes the main domain-specific factors and their simulation implications.

\subsection{Implications for Simulation}

A planetary robotics simulator should support the task-relevant physical,
sensing, and operational assumptions behind a study, not only render planetary
scenes. These domain factors motivate four considerations for
perception-oriented simulation. First, the simulator should represent terrain and
robot--environment interaction at the level required by the target task
\cite{Karl2004,Hockman2017,Helmick2009}. Second, it should reproduce sensing
conditions that matter for localization or scene interpretation, including low
texture and extreme illumination
\cite{Strader2020,Geromichalos2020,Furgale2010,Meyer2021}. Third, it should
make the operational assumptions of the target autonomy stack explicit,
including software integration, communication delay, onboard computation, and
energy budgets \cite{FIngrand2007,Gu2018,Gerdes2020}. Fourth, it should account
for the diversity of planetary robotic platforms, including rovers, legged
robots, aerial vehicles, and multi-robot systems
\cite{PArm2023,Lutz2020,Schuster2019}.

These considerations motivate the survey axes in Section~\ref{sec:survey}.
Because representative planetary testing is difficult to reproduce, we also
treat openness and availability as a practical axis for reuse and comparison,
as illustrated in Fig.~\ref{fig:overview}.

\section{Survey of Existing Simulation Works}
\label{sec:survey}

\subsection{Taxonomy and analysis criteria}

Building on Section~\ref{sec:prelim}, we organize recent planetary simulation
works along four analysis axes: openness and availability, scenario and
platform coverage, sensor and perception support, and environmental and
operational realism. The surveyed set is intentionally bounded in scope: we
focus on works from 2020 onward with publicly accessible
scholarly documentation that explicitly present planetary simulation, rendering,
synthetic-data generation, or benchmark works relevant to robotic perception
and autonomy. We therefore exclude general-purpose engines by themselves,
internal tools, challenge-only environments, and projects without accompanying
scholarly documentation. To identify candidate works within this scope, we
searched arXiv, IEEE Xplore, and OpenReview using combinations of search terms
such as ``lunar,'' ``mars,'' ``simulation,'' ``navigation,'' and ``simulator.''

We analyze these entries as a heterogeneous set of works, not as directly
comparable complete simulators. Here, rover-relevant denotes works that are
not full rover simulators but still provide terrain, stereo/depth, planning, or
perception inputs useful for rover autonomy studies. Table~\ref{tab:sim-profile}
summarizes the resulting twenty-four works.\footnote{Paper and simulation-resource links were checked on September 19, 2026.}
The following distributions
should be read as descriptive patterns within this set, not as field-wide
prevalence estimates.

\begin{table*}[!tp]
\centering
\hypersetup{hidelinks}
\caption{Overview of the twenty-four surveyed works. Access is \emph{paper / simulation resources}: P = public publication page or full-text link; C = work-specific code; C* = partial components or evaluation code; B = simulator executable; D = data/assets; R = request channel; -- = no release identified. R1/R2 = ROS 1/2; ROS = supported, version unspecified. St = Stereo; Sem = Semantic; Seg = Segmentation; Enc = Wheel encoder; Alt = Altimeter; Inc = Inclinometer.}
\label{tab:sim-profile}
\resizebox{\textwidth}{!}{%
\scriptsize
\setlength{\tabcolsep}{5pt}
\renewcommand{\arraystretch}{1.10}
\begin{tabular}{lcccccccl}
\hline
\textbf{Work} & \textbf{Year} & \textbf{Domain} & \textbf{Category} & \textbf{Engine / Tool} & \textbf{ROS} & \textbf{Access} & \textbf{Terrain Source} & \textbf{Sensors / Outputs} \\
\hline
Giubilato et al.~\cite{Giubilato2020} & 2020 & Mars & Rover & Gazebo & ROS & \href{https://arxiv.org/abs/2006.00057}{P} / \href{https://github.com/MorpheusPD/MarsSim/tree/16e141cfc821aa283ff51a25388c4191bdbb8883}{C} & Real & St Cam, 3D LiDAR \\
Toupet et al.~\cite{Toupet2020ENavSim} & 2020 & Mars & Rover & ROS-based & ROS & \href{https://josh.vanderhook.info/media/pdf/aero2020roversim.pdf}{P} / -- & Simulated & St Cam, rover state \\
OAISYS~\cite{Mueller2021} & 2021 & Planet. & Multi-platform & Blender & -- & \href{https://elib.dlr.de/145997/1/224_paper.pdf}{P} / \href{https://github.com/DLR-RM/oaisys/tree/febd3f6e1700eb89d21071cfd15e9029f11fb5c8}{C} & Synthetic & RGB-D (extensible) \\
GraspPlanetary~\cite{GraspPlanetary2022} & 2022 & Lunar & Manipulation & Gazebo + DART & R2 & \href{https://arxiv.org/abs/2208.00818v1}{P} / \href{https://github.com/AndrejOrsula/drl_grasping/tree/b2283544d19c3b202b14caa03e233ac4598ccee2}{C} & Procedural & Depth/intensity, octrees \\
MarsSim~\cite{MarsSim2023} & 2023 & Mars & Rover & Gazebo & ROS & \href{https://doi.org/10.1109/TAES.2022.3207705}{P} / -- & Real & Mast-mounted Cam \\
SEELO~\cite{Cloud2023} & 2023 & Lunar & Excavator & Unity & -- & \href{https://ntrs.nasa.gov/api/citations/20220014197/downloads/root_v2.pdf}{P} / -- & Real & Vision, IMU \\
LunarSim~\cite{LunarSim2023} & 2023 & Lunar & Rover & Unity & R2 & \href{https://mdpi-res.com/d_attachment/applsci/applsci-13-12401/article_deploy/applsci-13-12401.pdf}{P} / \href{https://github.com/PUTvision/LunarSim/releases/tag/v0.1.0}{B} & Synthetic & RGB-D, IMU \\
Jiang et al.~\cite{Yunn2023} & 2023 & Mars & Rover & Blender & -- & \href{https://doi.org/10.1109/ONCON60463.2023.10431105}{P} / -- & Synthetic & RGB-D + Sem \\
DUST~\cite{DUST2023} & 2023 & Lunar & Rover-relevant & Unreal Engine & -- & \href{https://ntrs.nasa.gov/api/citations/20220014707/downloads/DUST_IEEE2023Paper.pdf}{P} / \href{https://software.nasa.gov/software/MSC-27522-1}{R} & Real/DEM & Rendered terrain views \\
OmniLRS~\cite{Richard_2024} & 2024 & Lunar & Rover & Isaac Sim & R1/R2 & \href{https://arxiv.org/abs/2309.08997v1}{P} / \href{https://github.com/OmniLRS/OmniLRS/tree/5429512dfb80808a047e44c1b97dc0aadc3ce327}{C} & Real + proc. & RGB-D, 2D/3D LiDAR, IMU \\
MarsDrone~\cite{MarsDrone2024} & 2024 & Mars & Aerial & Unreal Engine & -- & \href{https://doi.org/10.1109/AERO58975.2024.10521330}{P} / -- & Real & Cam, Alt, Inc, IMU \\
RLRoverLab~\cite{RLRoverLAB2024} & 2024 & Lu./Ma. & Rover & Isaac Sim & ROS & \href{https://doi.org/10.1109/iSpaRo60631.2024.10687686}{P} / \href{https://github.com/abmoRobotics/RLRoverLab/tree/578906ad5c3383087aeec038be82b87e05d4ee2c}{C} & Synthetic & RGB Cam \\
LunarRoverSim~\cite{LunarRoverSim2024} & 2024 & Lunar & Rover & PyBullet & -- & \href{https://doi.org/10.1109/CACRE62362.2024.10634867}{P} / \href{https://github.com/assawayut/LunarRoverSim/tree/896964558c4e98599948cefd6afcb0ba08adcb80}{B} & Config. lunar & RGB-D + Seg \\
ISMRS~\cite{ISMRS2024} & 2024 & Mars & Rover & Isaac Sim & R2 & \href{https://doi.org/10.21203/rs.3.rs-4851864/v1}{P} / -- & Real & St Cam, IMU, LiDAR, Enc \\
GUISS~\cite{GUISS2024} & 2024 & Icy moon & Rover-relevant & Blender-based & -- & \href{https://arxiv.org/abs/2401.12414v1}{P} / \href{https://github.com/nasa-jpl/guiss/tree/932c061a963707ab98364209b8e71daa82a2cff3}{C} & Synthetic & Stereo imagery \\
Batagoda et al.~\cite{Chrono2024} & 2025 & Lunar & Excavator & Chrono & R2 & \href{https://arxiv.org/abs/2410.04371v1}{P} / \href{https://github.com/projectchrono/chrono/tree/a92c6f72f422fbcafe0b37125d4070cb6a3b5803/src/chrono_sensor/optix/shaders/camera_hapke_shader.cuh}{C*} & Real (Earth) & Cam, sensor sim. \\
LuSeg/LESS~\cite{LuSeg2025} & 2025 & Lunar & Rover & UE + AirSim & ROS & \href{https://arxiv.org/abs/2503.11409v1}{P} / \href{https://github.com/nubot-nudt/LuSeg/tree/0f64efed77a08aa9f873a21e1c3098903d005baf}{C*} & Procedural & RGB-D, LiDAR, IMU, Seg \\
Linde et al.~\cite{Mattias2025} & 2025 & Lunar & Exc./Truck & AGX Dynamics & R2 & \href{https://arxiv.org/abs/2505.22091v2}{P} / -- & Synthetic & Observation sensor \\
Lindmark et al.~\cite{Daniel2025} & 2025 & Lunar & Rover & OpenPLX + AGX & R2 & \href{https://arxiv.org/abs/2509.12367v1}{P} / -- & LDEM import (planned) & RGB-D, LiDAR, IMU \\
PlanetaryPathBench~\cite{PlanetaryPathBench2025} & 2025 & Ma./Lu. & Rover-relevant & Benchmark maps & -- & \href{https://arxiv.org/abs/2512.21438v1}{P} / \href{https://github.com/mchancan/PlanetaryPathBench/tree/86dc4b63f14551e55607bad1ce520164d8a96359}{C} & Real/DEM & Occupancy maps \\
SRB~\cite{SRB2025} & 2025 & Planet. & Multi-platform & Isaac Sim/Lab & R2 & \href{https://arxiv.org/abs/2509.23328v1}{P} / \href{https://github.com/AndrejOrsula/space_robotics_bench/tree/7528ff81f1ac0b34ba259b1ae150fdb4f6a90b5e}{C} & Procedural & RGB-D, IMU, Seg, state \\
MARTIAN~\cite{MARTIAN2026} & 2026 & Mars & Aerial & Blender & -- & \href{https://arxiv.org/abs/2605.29647v1}{P} / \href{https://github.com/nasa-jpl/martian/tree/f03c16d47b42ebabb3e63d392576ea75b8bf5b87}{C} & Real map & Aerial RGB + pose \\
POLAR-Sim~\cite{POLARSim2026} & 2026 & Lunar & Rover & Chrono & -- & \href{https://arxiv.org/abs/2309.12397v2}{P} / \href{https://github.com/uwsbel/POLAR-Sim/tree/593ed65c901bb53a05dc0a7363f5192d1171b85d}{C*/D} & Real + synth. & RGB, annotations, terrain meshes \\
Kurt et al.~\cite{DualSim2026} & 2026 & Lunar & Rover & MuJoCo + Isaac & -- & \href{https://openreview.net/forum?id=YdtCNcDCIV}{P} / -- & Synthetic & RGB-D + pose \\
\hline
\end{tabular}%
}
\vspace{-8pt}
\end{table*}

\subsection{Openness and Availability}
\label{sec:openness}

\begin{figure}[!t]
\centering
\begin{tikzpicture}[
    bar/.style={fill=#1, minimum height=0.35cm, anchor=west, inner sep=0pt},
    lbl/.style={font=\sffamily\scriptsize, anchor=east},
    val/.style={font=\sffamily\scriptsize\bfseries, anchor=west},
]
\node[font=\sffamily\small\bfseries, anchor=west] at (0, 1.6) {(a) Online Availability};
\node[bar=colCardTeal, minimum width=3.50cm] at (2.6, 1.05) (b1) {};
\node[bar=colCardAmber, minimum width=0.25cm] at (2.6, 0.55) (b2) {};
\node[bar=colCardTerra, minimum width=2.25cm] at (2.6, 0.05) (b3) {};
\node[lbl] at (2.5, 1.05) {Public releases (14)};
\node[lbl] at (2.5, 0.55) {Request (1)};
\node[lbl] at (2.5, 0.05) {Not identified (9)};
\node[val, color=colCardTeal!80!black] at (b1.east) {\;58.3\%};
\node[val, color=colCardAmber!70!black] at (b2.east) {\;4.2\%};
\node[val, color=colCardTerra!80!black] at (b3.east) {\;37.5\%};
\draw[black!25] (2.6, -0.2) -- (2.6, 1.3);

\node[font=\sffamily\small\bfseries, anchor=west] at (0, -0.55) {(b) Platform Category};
\node[bar=colCardIndigo, minimum width=4.44cm] at (2.6, -1.10) (p1) {};
\node[bar=colCardAmber, minimum width=0.83cm] at (2.6, -1.60) (p2) {};
\node[bar=colCardTeal, minimum width=0.56cm] at (2.6, -2.10) (p3) {};
\node[bar=colCardTerra, minimum width=0.56cm] at (2.6, -2.60) (p4) {};
\node[lbl] at (2.5, -1.10) {Rover-relevant (16)};
\node[lbl] at (2.5, -1.60) {Construction (3)};
\node[lbl] at (2.5, -2.10) {Aerial (2)};
\node[lbl] at (2.5, -2.60) {Multi-platform (2)};
\node[bar=colCardAmber, minimum width=0.28cm] at (2.6, -3.10) (p5) {};
\node[lbl] at (2.5, -3.10) {Manipulation (1)};
\node[val, color=colCardAmber!70!black] at (p5.east) {\;4.2\%};
\node[val, color=colCardIndigo!80!black] at (p1.east) {\;66.7\%};
\node[val, color=colCardAmber!70!black] at (p2.east) {\;12.5\%};
\node[val, color=colCardTeal!80!black] at (p3.east) {\;8.3\%};
\node[val, color=colCardTerra!80!black] at (p4.east) {\;8.3\%};
\draw[black!25] (2.6, -3.35) -- (2.6, -0.85);

\end{tikzpicture}
\caption{Distribution of surveyed simulation works by (a) access to work-specific simulation resources and (b) primary platform or task relevance. Public releases include code, partial components, executables, and data (Table~\ref{tab:sim-profile}).}
\label{fig:distributions}
\vspace{-8pt}
\end{figure}
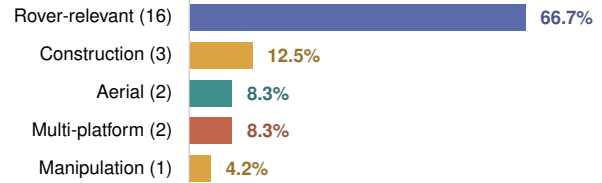

Simulation resources are publicly available for fourteen of the twenty-four
surveyed works (58.3\%), while DUST~\cite{DUST2023} offers a request channel
(4.2\%) and no corresponding release was identified for nine works (37.5\%)
(Fig.~\ref{fig:distributions}a). All surveyed works have public publication pages or full-text links,
provided in Table~\ref{tab:sim-profile}; simulation-resource availability is
reported separately.

\subsection{Scenario and Platform Coverage}
\label{sec:scenario}

Rover scenarios and rover-relevant terrain or planning tasks form the largest
category, accounting for sixteen of the twenty-four surveyed works
(Fig.~\ref{fig:distributions}b). The remaining platform and task categories
cover construction, aerial platforms, manipulation, and multi-platform
applications. Representative examples include SEELO for lunar excavation
\cite{Cloud2023}, MarsDrone for aerial simulation \cite{MarsDrone2024},
GraspPlanetary for lunar manipulation \cite{GraspPlanetary2022}, and SRB for
tasks involving multiple robot embodiments \cite{SRB2025}.

\subsection{Sensor and Perception Support}
\label{sec:sensor}

Sensor and perception support is one of the clearest convergence points
across the surveyed literature. Representative examples include multi-modal
synthetic data generation, computer-vision prototyping, segmentation-oriented
lunar datasets, map-based localization imagery, and vision-oriented rover
navigation \cite{Richard_2024,LunarSim2023,LuSeg2025,MARTIAN2026,DualSim2026}.

Table~\ref{tab:sensor-gap} contrasts the sensing modalities
emphasized in current simulation works against those discussed in
planetary mission documentation and navigation surveys.
The dominant supported modalities are cameras, RGB-D, LiDAR, and IMUs,
together with perception-oriented outputs such as semantic labels, depth, and annotations.
However, explicit support for more specialized modalities
(ground-penetrating radar, thermal, event-based, celestial-navigation sensors)
is reported much less often.
Table~\ref{tab:sim-profile} further details the terrain data source and the
supported sensor modalities and outputs per work.

\begin{table}[!t]
\centering
\caption{Sensing modality coverage reported in the surveyed works and in mission, planned-mission, prototype-stage, or agency research-stage literature. Entries indicate reported support or use rather than comprehensive coverage of the field or baseline simulator requirements.}
\label{tab:sensor-gap}
\resizebox{\columnwidth}{!}{%
\small
\setlength{\tabcolsep}{4pt}
\begin{tabular}{l c c}
\toprule
\textbf{Sensing Modality} & \textbf{Surveyed Works} & \textbf{Mission / Prototype} \\
\midrule
Mono / Stereo Camera & \yes & \yes~\cite{bluethmann2024viper} \\
RGB-D & \yes & \yes~\cite{vanderMeer2023REALMS} \\
2D / 3D LiDAR & \yes & \yes~\cite{Gu2018} \\
IMU & \yes & \yes~\cite{bluethmann2024viper} \\
Wheel Encoder / Odometry & \yes~\cite{ISMRS2024,Daniel2025,Cloud2023} & \yes~\cite{Gu2018} \\
Semantic / Instance Labels & \yes & -- \\
\midrule
Star Tracker / Celestial Nav. & \no & \yes~\cite{mckee2025survey} \\
Ground-Penetrating Radar & \no & \yes~\cite{RIMFAXNASA2020}, \cite{CADREMission}, \cite{Sheppard2025MarsLGPR} \\
Thermal / IR Sensor & \no & \yes~\cite{NASAMEDATIRS}, \cite{DLRMMXMiniRAD} \\
Event-based Camera & \no & \pmark~\cite{ESAACTEventVision} \\
ToF / Flash LiDAR & \yes~\cite{MarsDrone2024} & \pmark~\cite{NASATechPortFlashLidar}, \cite{NASASPLICEHDL} \\
\bottomrule
\end{tabular}%
}
\vspace{1pt}
\par\noindent{\scriptsize \yes{} = reported support or use; \no{} = not reported in the reviewed descriptions; \pmark{} = limited or indirect support; entries are based on how each work reports its capabilities.}
\par\noindent{\scriptsize Reported support denotes capabilities implemented and exposed by the work itself; engine-level possibilities that require substantial user-side implementation are not counted. For ToF, this refers to the simulated altimeter in MarsDrone; Flash LiDAR imaging is not established by this example.}
\vspace{-8pt}
\end{table}

\subsection{Environmental and Operational Realism}
\label{sec:realism}

Environmental and operational realism are treated here as complementary aspects
of planetary simulation. The former includes terrain geometry, regolith contact,
illumination, gravity, and sensing conditions, while the latter concerns
communication delay, onboard computation, energy use, and autonomy-stack
execution. Fig.~\ref{fig:realism-scene} summarizes where these factors appear in
a mission scene and how they are represented across the surveyed works.

The surveyed works address environmental realism through terrain and contact
modeling, physics-based sensor simulation, and task-specific image generation
\cite{Chrono2024,POLARSim2026,GUISS2024,MARTIAN2026,DualSim2026}.
Related studies on OmniLRS further develop terrain deformation and
wheel--regolith interaction models \cite{Junnosuke2024,Jakob2026}.

Reported operational treatments vary in scope (Table~\ref{tab:operational}).
Energy-related treatments range from battery depletion and power consumption
in SEELO to actuator-work estimates in Linde et al. and task-dependent
penalties in SRB \cite{Cloud2023,Mattias2025,SRB2025}.
LunarRoverSim additionally models motor electrical losses and
temperature-dependent torque limits \cite{LunarRoverSim2024}.
Communication and onboard computation receive more limited treatment,
exemplified by DUST's line-of-sight visualization and LunarSim's
embedded-inference experiments \cite{DUST2023,LunarSim2023}.

\begin{table}[!t]
\centering
\caption{Reported treatment of communication, onboard computational resource limits, and energy in the surveyed works.}
\label{tab:operational}
\scriptsize
\setlength{\tabcolsep}{4pt}
\renewcommand{\arraystretch}{1.04}
\begin{subtable}{\columnwidth}
\centering
\caption{Coverage across the surveyed works.}
\label{tab:operational-coverage}
\begin{tabular*}{\linewidth}{@{\extracolsep{\fill}}lccc@{}}
\toprule
\textbf{Work} & \textbf{Comm.} & \textbf{Compute} & \textbf{Energy} \\
\midrule
Giubilato et al.~\cite{Giubilato2020} & $\times$ & $\times$ & $\times$ \\
Toupet et al.~\cite{Toupet2020ENavSim} & $\times$ & $\times$ & $\times$ \\
OAISYS~\cite{Mueller2021} & $\times$ & $\times$ & $\times$ \\
GraspPlanetary~\cite{GraspPlanetary2022} & $\times$ & $\times$ & $\times$ \\
MarsSim~\cite{MarsSim2023} & $\times$ & $\times$ & $\times$ \\
SEELO~\cite{Cloud2023} & $\times$ & $\times$ & $\bigcirc$ \\
LunarSim~\cite{LunarSim2023} & $\times$ & $\triangle$ & $\times$ \\
Jiang et al.~\cite{Yunn2023} & $\times$ & $\times$ & $\times$ \\
DUST~\cite{DUST2023} & $\triangle$ & $\times$ & $\times$ \\
OmniLRS~\cite{Richard_2024} & $\times$ & $\times$ & $\times$ \\
MarsDrone~\cite{MarsDrone2024} & $\times$ & $\times$ & $\times$ \\
RLRoverLab~\cite{RLRoverLAB2024} & $\times$ & $\times$ & $\times$ \\
LunarRoverSim~\cite{LunarRoverSim2024} & $\times$ & $\times$ & $\bigcirc$ \\
ISMRS~\cite{ISMRS2024} & $\times$ & $\times$ & $\times$ \\
GUISS~\cite{GUISS2024} & $\times$ & $\times$ & $\times$ \\
Batagoda et al.~\cite{Chrono2024} & $\times$ & $\times$ & $\times$ \\
LuSeg/LESS~\cite{LuSeg2025} & $\times$ & $\times$ & $\times$ \\
Linde et al.~\cite{Mattias2025} & $\times^{\dagger}$ & $\times$ & $\bigcirc$ \\
Lindmark et al.~\cite{Daniel2025} & $\times$ & $\times$ & $\triangle$ \\
PlanetaryPathBench~\cite{PlanetaryPathBench2025} & $\times$ & $\times$ & $\times$ \\
SRB~\cite{SRB2025} & $\times$ & $\times$ & $\triangle$ \\
MARTIAN~\cite{MARTIAN2026} & $\times$ & $\times$ & $\times$ \\
POLAR-Sim~\cite{POLARSim2026} & $\times$ & $\times$ & $\times$ \\
Kurt et al.~\cite{DualSim2026} & $\times$ & $\times$ & $\times$ \\
\midrule
\textbf{$\bigcirc$ / $\triangle$ counts (24 reviewed)} & 0/1 & 0/1 & 3/2 \\
\bottomrule
\end{tabular*}
\par\smallskip\noindent $\bigcirc$ = explicit model or evaluation; $\triangle$ = limited or indirect treatment; $\times$ = no model or evaluation reported; $\dagger$ = future work only.
\end{subtable}
\par\medskip
\begin{subtable}{\columnwidth}
\centering
\caption{Reported treatments.}
\label{tab:operational-treatments}
\begin{tabularx}{\linewidth}{@{}l>{\raggedright\arraybackslash}X@{}}
\toprule
\textbf{Work} & \textbf{Reported treatment} \\
\midrule
SEELO & Battery depletion and power consumption \\
LunarSim & Embedded-inference evaluation \\
DUST & Line-of-sight visualization \\
LunarRoverSim & Motor electrical losses and thermal limits \\
Linde et al. & Actuator-work estimates \\
Lindmark et al. & Recap of Linde's construction study \\
SRB & Task-level energy/propellant penalties \\
\bottomrule
\end{tabularx}
\end{subtable}
\vspace{-8pt}
\end{table}



\definecolor{colCardTerra}{HTML}{C1664B}   
\definecolor{colCardAmber}{HTML}{D9A441}   
\definecolor{colCardIndigo}{HTML}{5C6BAB}  
\definecolor{colCardTeal}{HTML}{3E8E8B}    

\newcommand{\scardpng}[8]{%
    \fill[white, rounded corners=3pt] (#1,#2) rectangle (#1+#3,#2-#4);
    \draw[rounded corners=3pt, line width=0.7pt, draw=#5!55]
        (#1,#2) rectangle (#1+#3,#2-#4);
    \fill[#5] (#1+0.12,#2-#4) rectangle (#1+0.26,#2);
    \fill[#5] (#1+0.60,#2-0.38) circle (0.22);
    \node[font=\sffamily\scriptsize\bfseries, text=white, inner sep=0pt]
        at (#1+0.60,#2-0.38) {#6};
    \node[font=\sffamily\footnotesize, text=black!82, anchor=north west,
        align=left, text width=#3cm-1.00cm, inner sep=1pt]
        at (#1+0.92,#2-0.18) {\textbf{#7}\\[4.0pt]
        #8};
}

\begin{figure*}[!tp]
\centering
\resizebox{\textwidth}{!}{%
\begin{tikzpicture}[
    x=1cm, y=1cm,
    every node/.style={inner sep=1pt},
]

\node[anchor=south west, inner sep=0pt] at (0.15,4.60)
    {\includegraphics[width=24.30cm]{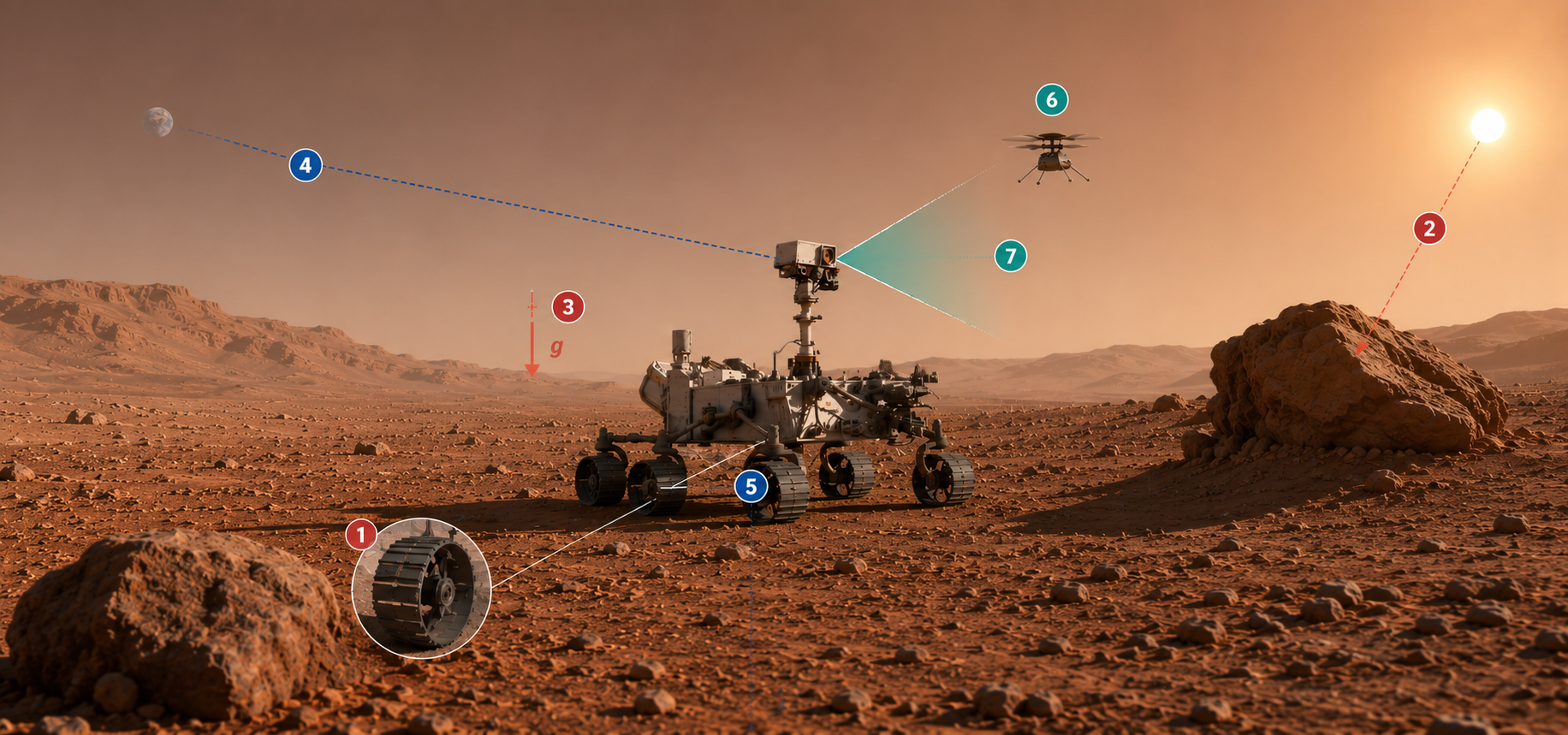}};

\scardpng{0.15}{4.45}{5.85}{2.05}{colCardTerra}{1}
    {Terrain Interaction}
    {\textbf{Implication:} rover mobility on deformable regolith\\[1.0pt]\textbf{Gap:} contact models are task-specific.}
\scardpng{6.30}{4.45}{5.85}{2.05}{colCardTerra}{2}
    {Illumination \& Visual}
    {\textbf{Implication:} illumination shapes perception\\[1.0pt]\textbf{Gap:} rendering remains task-specific.}
\scardpng{12.45}{4.45}{5.85}{2.05}{colCardTerra}{3}
    {Gravity \& Contact Regime}
    {\textbf{Implication:} low-g contact affects dynamics\\[1.0pt]\textbf{Gap:} assumptions are unevenly reported.}
\scardpng{18.60}{4.45}{5.85}{2.05}{colCardIndigo}{4}
    {Communication}
    {\textbf{Implication:} Earth--robot link delay affects autonomy\\[1.0pt]\textbf{Gap:} coverage models need qualification.}

\scardpng{0.15}{2.10}{5.85}{2.05}{colCardIndigo}{5}
    {Onboard Resources}
    {\textbf{Implication:} onboard compute and power bound execution\\[1.0pt]\textbf{Gap:} resource models differ in scope.}
\scardpng{6.30}{2.10}{5.85}{2.05}{colCardTeal}{6}
    {Platform Diversity}
    {\textbf{Implication:} rover-centered scenes limit mission scope\\[1.0pt]\textbf{Gap:} fewer heterogeneous missions.}
\scardpng{12.45}{2.10}{5.85}{2.05}{colCardTeal}{7}
    {Sensor \& Autonomy Stack}
    {\textbf{Implication:} sensing assumptions shape autonomy studies\\[1.0pt]\textbf{Gap:} specialized modalities are sparse.}

\end{tikzpicture}%
}%
\caption[Mission-scene view of environmental and operational realism in planetary simulation.]{Mission-scene view of how environmental and operational realism
factors appear in planetary simulation. The upper image provides the mission
context\protect\footnotemark, while the indexed blocks give an at-a-glance summary of the implication
and reporting gap associated with each factor. Table~\ref{tab:operational} provides the source-based operational comparison
and distinguishes explicit models from limited treatments.}
\label{fig:realism-scene}
\vspace{-8pt}
\end{figure*}

\section{Open Challenges and Future Directions}
\label{sec:challenges}

The four analysis axes of Section~\ref{sec:survey} reveal corresponding open challenges,
as traced in Fig.~\ref{fig:overview}.
This section discusses these challenges and concludes with a unifying direction.

\subsection{Limited openness and reusable public ecosystems}
\label{sec:ch-open}

As summarized in Section~\ref{sec:openness}, public availability remains uneven
across the surveyed works. For planetary robotics, this is especially relevant because
representative field testing and real mission data are limited, expensive, and
difficult to reproduce.
When a simulation work is publicly accessible, later studies can reuse its assets,
compare methods under shared settings, and reduce repeated engineering.
Future work could support broader reuse by sharing study-specific simulation
code and assets together with experimental configurations and runnable examples.
These resources would help other researchers recreate experimental setups and
compare methods under shared conditions.

\subsection{Rover concentration and limited scenario diversity}
\label{sec:ch-rover}

The surveyed literature remains rover-centered, but the broader planetary robotics
mission space already includes cooperative exploration \cite{PArm2023}, \cite{RDomingues2025},
aerial vehicles \cite{Lutz2020}, and multi-robot operation \cite{Schuster2019}.
NASA reporting has shown that Ingenuity provided scouting support
for Perseverance, illustrating the relevance of heterogeneous
aerial--ground cooperation \cite{jpl2022}.
Similarly, CADRE is designed around a team of small lunar rovers that conduct
autonomous distributed measurements, including surface imaging and subsurface
mapping with ground-penetrating radar \cite{CADREMission}, while CRESCENT
addresses constrained lunar rover planning for the same multi-agent mission
\cite{Cauligi2025CRESCENT}.
These examples show that the gap is broader than platform coverage alone.
Reusable scenarios remain limited for heterogeneous, cooperative, and
science-driven operations such as planetary skylight and lava-cave
exploration~\cite{RDomingues2025}.
Future planetary simulation work would therefore benefit from broader scenario
coverage aligned with such mission concepts and recent fielded or planned
robotic systems.

\subsection{Insufficient support for diverse sensing modalities}
\label{sec:ch-sensor}

As shown in Table~\ref{tab:sensor-gap}, the sensing assumptions in the
surveyed simulation works are narrower than the broader sensing possibilities
discussed in planetary navigation and mission documentation.
The relevant issue is not whether every simulation work should implement every
mission payload, but whether task-relevant sensing assumptions can be studied
when they matter for localization, traversability estimation, or autonomy.
Ground-penetrating radar illustrates this distinction. RIMFAX on Perseverance
and the planned CADRE mission show the mission relevance of subsurface radar
\cite{RIMFAXNASA2020}, \cite{CADREMission}, while recent work~\cite{Sheppard2025MarsLGPR}
estimates relative displacement from sequential GPR observations and
integrates it into rover state estimation through an extended Kalman filter,
expanding its role from scientific sounding to autonomy-oriented state estimation.
Other navigation-oriented modalities point to the same issue. Star trackers
and celestial-navigation sensors are discussed in lunar surface navigation
surveys \cite{mckee2025survey}.
Thermal/IR sensing, event-based vision, and ToF/Flash LiDAR provide additional
examples from mission, prototype, or agency research-stage contexts
\cite{NASAMEDATIRS}, \cite{ESAACTEventVision}, \cite{NASATechPortFlashLidar}.
Broader simulation support for these modalities would expand the range of
algorithmic studies that can be performed before field deployment, including
subsurface-aided localization, thermal traversability assessment, event-based
perception, and Flash LiDAR hazard detection. This would make planetary
simulation a stronger tool for developing and stress-testing perception and
autonomy methods under mission-relevant sensing assumptions.

\subsection{Limited treatment of operational constraints}
\label{sec:ch-ops}

VIPER documentation reports power and communication
considerations~\cite{bluethmann2024viper} and round-trip communication
delays~\cite{fong2022viper}.
The lunar surface navigation survey~\cite{mckee2025survey} identifies computational cost and hardware requirements
for several navigation modes.
Operational constraints also affect the execution of autonomy pipelines.
The Mars 2020 autonomy system overlaps image processing, navigation processing,
and driving during a traverse \cite{McHenry2020Mars2020AutoNav}, illustrating
the relevance of execution timing to rover operation.
Some surveyed works provide limited treatments of operational constraints,
including line-of-sight visualization, embedded-inference evaluation, and
task-level energy analysis (Table~\ref{tab:operational}).
Gaps remain in how these treatments account for mission-wide resource budgets
and their effects on autonomy-stack execution.
Future work could help address these gaps through more explicit modeling and
evaluation of communication, onboard computation, and energy constraints.
\footnotetext{The Mars scene image in Fig.~\ref{fig:realism-scene} was generated using ChatGPT Image 2.0.}

\subsection{Toward perception-oriented planetary simulation ecosystems}

Addressing these challenges calls for reusable simulation resources that support
diverse robotic platforms and sensing modalities while making environmental
and operational assumptions explicit. Shared task definitions, configurable
scenarios, and clearly specified evaluation procedures could help researchers
reuse these resources and compare methods under comparable experimental
conditions.

\section{Practical Recommendations}
\label{sec:ready}

For researchers entering planetary robotics simulation, the surveyed landscape
can be condensed into a small set of practical starting points, selected from
works with publicly available simulation resources (\S\,\ref{sec:openness}) and capabilities
suited to the following tasks.
For lunar rover studies, OmniLRS~\cite{Richard_2024} is a practical entry point: it
couples photorealistic Isaac Sim rendering with ROS\,1/ROS\,2 interfaces,
multi-robot support, and multi-modal synthetic data generation.
LunarSim~\cite{LunarSim2023} supports ROS\,2-based computer-vision prototyping,
while RLRoverLab~\cite{RLRoverLAB2024} targets
learning-based rover autonomy on both lunar and Martian terrain.
For lunar grasping, GraspPlanetary~\cite{GraspPlanetary2022} provides a
Gazebo/ROS\,2 learning environment; SRB~\cite{SRB2025} supports a broader
range of planetary robot-learning tasks on Isaac Sim/Lab.
For Mars rover work in a classic ROS/Gazebo stack, the framework of
Giubilato et al.~\cite{Giubilato2020} provides a public simulation example.
When terramechanics and physical fidelity dominate the research question,
Chrono-based works provide relevant starting points, including Batagoda et al.~\cite{Chrono2024}
and POLAR-Sim~\cite{POLARSim2026}, the latter adding lunar lighting realism
and a rover digital twin.
For synthetic data generation beyond a single platform,
OAISYS~\cite{Mueller2021} provides a Blender-based pipeline with semantic
metadata, and MARTIAN~\cite{MARTIAN2026} renders Mars aerial imagery for
map-based localization studies.
Finally, PlanetaryPathBench~\cite{PlanetaryPathBench2025} offers
terrain-derived benchmark maps for evaluating rover path planning without a
full simulation stack.
These recommendations are intended to help first-time users identify suitable
starting points based on their research goals and required simulation
capabilities.

\section{Conclusion}
\label{sec:conclusion}

This paper surveyed twenty-four simulation works for planetary robotic perception
and autonomy, comparing their availability, platform coverage, sensing support,
and environmental and operational realism. The surveyed works report support
for visual rendering, synthetic-data generation, and task-specific physical
simulation. Within this set, public availability is uneven, rover-relevant
applications dominate, specialized sensing is sparsely reported, and operational
constraints are described at varying levels of detail. These findings motivate
broader sharing of reusable simulation resources and clearer reporting of
task-specific assumptions to support reproducible perception and autonomy
research.


\bibliographystyle{IEEEtran}
{\small
\bibliography{ref}

\begin{thebibliography}{10}
\providecommand{\url}[1]{#1}
\csname url@samestyle\endcsname
\providecommand{\newblock}{\relax}
\providecommand{\bibinfo}[2]{#2}
\providecommand{\BIBentrySTDinterwordspacing}{\spaceskip=0pt\relax}
\providecommand{\BIBentryALTinterwordstretchfactor}{4}
\providecommand{\BIBentryALTinterwordspacing}{\spaceskip=\fontdimen2\font plus
\BIBentryALTinterwordstretchfactor\fontdimen3\font minus
  \fontdimen4\font\relax}
\providecommand{\BIBforeignlanguage}[2]{{%
\expandafter\ifx\csname l@#1\endcsname\relax
\typeout{** WARNING: IEEEtran.bst: No hyphenation pattern has been}%
\typeout{** loaded for the language `#1'. Using the pattern for}%
\typeout{** the default language instead.}%
\else
\language=\csname l@#1\endcsname
\fi
#2}}
\providecommand{\BIBdecl}{\relax}
\BIBdecl

\bibitem{YGao2017}
Y.~Gao and S.~Chien, ``Review on space robotics: Toward top-level science
  through space exploration,'' \emph{Science Robotics}, vol.~2, no.~7, p.
  eaan5074, 2017.

\bibitem{FIngrand2007}
F.~Ingrand \emph{et~al.}, ``Decisional autonomy of planetary rovers,''
  \emph{Journal of Field Robotics}, vol.~24, no.~7, pp. 559--580, 2007.

\bibitem{PArm2023}
P.~Arm \emph{et~al.}, ``Scientific exploration of challenging planetary analog
  environments with a team of legged robots,'' \emph{Science Robotics}, vol.~8,
  no.~80, p. eade9548, 2023.

\bibitem{RDomingues2025}
R.~Domínguez \emph{et~al.}, ``Cooperative robotic exploration of a planetary
  skylight surface and lava cave,'' \emph{Science Robotics}, vol.~10, no. 105,
  p. eadj9699, 2025.

\bibitem{Strader2020}
J.~Strader \emph{et~al.}, ``Perception-aware autonomous mast motion planning
  for planetary exploration rovers,'' \emph{Journal of Field Robotics},
  vol.~37, no.~5, pp. 812--829, 2020.

\bibitem{Geromichalos2020}
D.~Geromichalos \emph{et~al.}, ``{SLAM} for autonomous planetary rovers with
  global localization,'' \emph{Journal of Field Robotics}, vol.~37, no.~5, pp.
  830--847, 2020.

\bibitem{Censible2024}
J.~Nash \emph{et~al.}, ``Censible: A robust and practical global localization
  framework for planetary surface missions,'' in \emph{2024 IEEE International
  Conference on Robotics and Automation (ICRA)}, 2024, pp. 8642--8648.

\bibitem{Mueller2021}
M.~G. Müller \emph{et~al.}, ``A photorealistic terrain simulation pipeline for
  unstructured outdoor environments,'' in \emph{2021 IEEE/RSJ International
  Conference on Intelligent Robots and Systems (IROS)}, 2021, pp. 9765--9772.

\bibitem{Richard_2024}
A.~Richard \emph{et~al.}, ``{OmniLRS}: A photorealistic simulator for lunar
  robotics,'' in \emph{2024 IEEE International Conference on Robotics and
  Automation (ICRA)}, 2024, pp. 16\,901--16\,907.

\bibitem{GraspPlanetary2022}
A.~Orsula, S.~Bøgh, M.~Olivares-Mendez, and C.~Martinez, ``Learning to grasp
  on the moon from 3d octree observations with deep reinforcement learning,''
  in \emph{2022 IEEE/RSJ International Conference on Intelligent Robots and
  Systems (IROS)}, 2022, pp. 4112--4119.

\bibitem{Cloud2023}
J.~M. Cloud \emph{et~al.}, ``Lunar excavator mission operations using dynamic
  movement primitives,'' in \emph{2023 IEEE/RSJ International Conference on
  Intelligent Robots and Systems (IROS)}, 2023, pp. 10\,708--10\,715.

\bibitem{Daniel2025}
D.~Lindmark \emph{et~al.}, ``An integrated process for design and control of
  lunar robotics using {AI} and simulation,'' arXiv preprint arXiv:2509.12367,
  2025.

\bibitem{POLARSim2026}
B.-H. Chen \emph{et~al.}, ``{POLAR-Sim}: Augmenting {NASA}'s {POLAR} dataset
  for data-driven lunar perception and rover simulation,'' \emph{IEEE Aerospace
  and Electronic Systems Magazine}, vol.~41, no.~1, pp. 4--18, 2026.

\bibitem{LunarSim2023}
D.~Pieczyński \emph{et~al.}, ``{LunarSim}: Lunar rover simulator focused on
  high visual fidelity and {ROS} 2 integration for advanced computer vision
  algorithm development,'' \emph{Applied Sciences}, vol.~13, no.~22, 2023, art.
  no. 12401.

\bibitem{Chrono2024}
N.~M. Batagoda \emph{et~al.}, ``A physics-based sensor simulation environment
  for lunar ground operations,'' in \emph{2025 IEEE Aerospace Conference},
  2025, pp. 1--20.

\bibitem{Mattias2025}
M.~Linde \emph{et~al.}, ``A simulation framework for autonomous lunar
  construction work,'' arXiv preprint arXiv:2505.22091, 2025.

\bibitem{DUST2023}
L.~Bingham \emph{et~al.}, ``Digital lunar exploration sites unreal simulation
  tool ({DUST}),'' in \emph{2023 IEEE Aerospace Conference}, 2023, pp. 1--12.

\bibitem{LunarRoverSim2024}
A.~Khunmaturod and D.~E. Chang, ``Development of a lunar rover simulator with
  an interface for reinforcement learning,'' in \emph{2024 9th International
  Conference on Automation, Control and Robotics Engineering (CACRE)}, 2024,
  pp. 117--123.

\bibitem{LuSeg2025}
S.~Jiao \emph{et~al.}, ``{LuSeg}: Efficient negative and positive obstacles
  segmentation via contrast-driven multi-modal feature fusion on the lunar,''
  in \emph{2025 IEEE/RSJ International Conference on Intelligent Robots and
  Systems (IROS)}, 2025, pp. 5962--5969.

\bibitem{DualSim2026}
U.~Kurt \emph{et~al.}, ``A dual-simulator framework for physics based
  locomotion and vision based navigation of planetary rovers,'' 2026.

\bibitem{RLRoverLAB2024}
A.~B. Mortensen and S.~Bøgh, ``{RLRoverLAB}: An advanced reinforcement
  learning suite for planetary rover simulation and training,'' in \emph{2024
  International Conference on Space Robotics (iSpaRo)}, 2024, pp. 273--277.

\bibitem{GUISS2024}
R.~Bhaskara \emph{et~al.}, ``Icy {Moon} surface simulation and stereo depth
  estimation for sampling autonomy,'' in \emph{2024 IEEE Aerospace
  Conference}.\hskip 1em plus 0.5em minus 0.4em\relax IEEE, Mar. 2024, p.
  1–16.

\bibitem{PlanetaryPathBench2025}
M.~Chancán \emph{et~al.}, ``Planetary terrain datasets and benchmarks for
  rover path planning,'' arXiv preprint arXiv:2512.21438, 2025.

\bibitem{SRB2025}
A.~Orsula, M.~Geist, M.~Olivares-Mendez, and C.~Martinez, ``{Space Robotics
  Bench: Robot Learning Beyond Earth},'' \emph{arXiv preprint
  arXiv:2509.23328}, 2025.

\bibitem{Giubilato2020}
R.~Giubilato \emph{et~al.}, ``Simulation framework for mobile robots in
  planetary-like environments,'' in \emph{2020 IEEE 7th International Workshop
  on Metrology for AeroSpace (MetroAeroSpace)}, 2020, pp. 594--599.

\bibitem{MarsSim2023}
R.~Zhou \emph{et~al.}, ``{MarsSim}: A high-fidelity physical and visual
  simulation for {Mars} rovers,'' \emph{IEEE Transactions on Aerospace and
  Electronic Systems}, vol.~59, no.~2, pp. 1879--1892, 2023.

\bibitem{MarsDrone2024}
G.~Zhang and Q.~Li, ``{MarsDrone}: A next-generation simulation system for
  {Mars} unmanned aerial vehicles,'' in \emph{2024 IEEE Aerospace Conference},
  2024, pp. 1--10.

\bibitem{MARTIAN2026}
D.~Pisanti and G.~Georgakis, ``{MARTIAN}: A rendering framework for aerial
  {Mars} imagery from {HiRISE} orbital data,'' arXiv preprint arXiv:2605.29647,
  2026.

\bibitem{Toupet2020ENavSim}
O.~Toupet \emph{et~al.}, ``A {ROS}-based simulator for testing the enhanced
  autonomous navigation of the {Mars} 2020 rover,'' in \emph{2020 IEEE
  Aerospace Conference}, 2020, pp. 1--11.

\bibitem{Yunn2023}
Y.~Jiang \emph{et~al.}, ``A {Mars} multi-terrain simulator using a modular
  terrain construction framework,'' in \emph{2023 IEEE 2nd Industrial
  Electronics Society Annual On-Line Conference (ONCON)}, 2023, pp. 1--6.

\bibitem{ISMRS2024}
W.~Liu \emph{et~al.}, ``High fidelity {Mars} rover simulation platform based on
  digital twin and machine learning integration: Innovation and application of
  {ISMRS},'' Research Square preprint, version 1, 2024.

\bibitem{Julio2023}
J.~A. Placed \emph{et~al.}, ``A survey on active simultaneous localization and
  mapping: State of the art and new frontiers,'' \emph{IEEE Transactions on
  Robotics}, vol.~39, no.~3, pp. 1686--1705, 2023.

\bibitem{Quoos2026}
M.~Quoos \emph{et~al.}, ``Survey on {AI}-enabled computer vision technologies
  and applications for space robotic missions,'' \emph{Journal of Field
  Robotics}, vol.~43, no.~4, pp. 2553--2584, 2026.

\bibitem{Lacroix2002}
S.~Lacroix \emph{et~al.}, ``Autonomous rover navigation on unknown terrains
  functions and integration,'' in \emph{Experimental Robotics VII}, D.~Rus and
  S.~Singh, Eds.\hskip 1em plus 0.5em minus 0.4em\relax Berlin, Heidelberg:
  Springer Berlin Heidelberg, 2001, pp. 501--510.

\bibitem{Ishigami2013}
G.~Ishigami \emph{et~al.}, ``Range-dependent terrain mapping and multipath
  planning using cylindrical coordinates for a planetary exploration rover,''
  \emph{Journal of Field Robotics}, vol.~30, no.~4, pp. 536--551, 2013.

\bibitem{Gonzalez2018}
R.~Gonzalez and K.~Iagnemma, ``Slippage estimation and compensation for
  planetary exploration rovers. state of the art and future challenges,''
  \emph{Journal of Field Robotics}, vol.~35, no.~4, pp. 564--577, 2018.

\bibitem{Karl2004}
K.~Iagnemma and S.~Dubowsky, ``Traction control of wheeled robotic vehicles in
  rough terrain with application to planetary rovers,'' \emph{The International
  Journal of Robotics Research}, vol.~23, no. 10-11, pp. 1029--1040, 2004.

\bibitem{Hockman2017}
B.~J. Hockman \emph{et~al.}, ``Design, control, and experimentation of
  internally-actuated rovers for the exploration of low-gravity planetary
  bodies,'' \emph{Journal of Field Robotics}, vol.~34, no.~1, pp. 5--24, 2017.

\bibitem{Helmick2009}
D.~Helmick \emph{et~al.}, ``Terrain adaptive navigation for planetary rovers,''
  \emph{Journal of Field Robotics}, vol.~26, no.~4, pp. 391--410, 2009.

\bibitem{Furgale2010}
P.~Furgale and T.~D. Barfoot, ``Visual teach and repeat for long-range rover
  autonomy,'' \emph{Journal of Field Robotics}, vol.~27, no.~5, pp. 534--560,
  2010.

\bibitem{Meyer2021}
L.~Meyer \emph{et~al.}, ``The {MADMAX} data set for visual-inertial rover
  navigation on {Mars},'' \emph{Journal of Field Robotics}, vol.~38, no.~6, pp.
  833--853, 2021.

\bibitem{Gu2018}
Y.~Gu \emph{et~al.}, ``Cataglyphis: An autonomous sample return rover,''
  \emph{Journal of Field Robotics}, vol.~35, no.~2, pp. 248--274, 2018.

\bibitem{Gerdes2020}
L.~Gerdes \emph{et~al.}, ``Efficient autonomous navigation for planetary rovers
  with limited resources,'' \emph{Journal of Field Robotics}, vol.~37, no.~7,
  pp. 1153--1170, 2020.

\bibitem{Lutz2020}
P.~Lutz \emph{et~al.}, ``{ARDEA}—an {MAV} with skills for future planetary
  missions,'' \emph{Journal of Field Robotics}, vol.~37, no.~4, pp. 515--551,
  2020.

\bibitem{Schuster2019}
M.~J. Schuster \emph{et~al.}, ``Distributed stereo vision-based {6D}
  localization and mapping for multi-robot teams,'' \emph{Journal of Field
  Robotics}, vol.~36, no.~2, pp. 305--332, 2019.

\bibitem{bluethmann2024viper}
B.~Bluethmann, ``{VIPER} rover overview and mobility and hardware,'' NASA
  Technical Reports Server (NTRS), Nov. 2024, presentation at the Technology
  Infusion Panel, Lunar Surface Innovation Consortium (LSIC), Las Vegas, NV,
  USA, November 13--15, 2024. Document ID: 20240013903.

\bibitem{vanderMeer2023REALMS}
D.~van~der Meer \emph{et~al.}, ``{REALMS}: Resilient exploration and lunar
  mapping system,'' \emph{Frontiers in Robotics and AI}, vol.~10, 2023, art.
  no. 1127496.

\bibitem{mckee2025survey}
P.~McKee, ``A survey of autonomous navigation techniques applicable to lunar
  surface exploration,'' Jan. 2025, {NASA} Technical Reports Server (NTRS),
  Document ID: 20250000993, Presentation, Report No. AAS 25-173, 47th Annual
  American Astronautical Society Guidance, Navigation and Control Conference,
  Breckenridge, CO, January 31--February 5, 2025.

\bibitem{RIMFAXNASA2020}
{NASA Science}, ``Searching for buried treasure on {Mars} with {RIMFAX},'' NASA
  Science, 2023, accessed: Sep. 16, 2026.

\bibitem{CADREMission}
{NASA Jet Propulsion Laboratory}, ``{CADRE},'' NASA Jet Propulsion Laboratory
  mission page, accessed: Sep. 16, 2026.

\bibitem{Sheppard2025MarsLGPR}
A.~Sheppard and K.~A. Skinner, ``{MarsLGPR}: {Mars} rover localization with
  ground penetrating radar,'' \emph{IEEE Transactions on Field Robotics},
  vol.~2, pp. 906--919, 2025.

\bibitem{NASAMEDATIRS}
J.~A. Rodriguez-Manfredi \emph{et~al.}, ``The {Mars Environmental Dynamics
  Analyzer}, {MEDA}. a suite of environmental sensors for the {Mars 2020}
  mission,'' \emph{Space Science Reviews}, vol. 217, no.~3, 2021, art. no. 48.

\bibitem{DLRMMXMiniRAD}
{German Aerospace Center (DLR)}, ``The {miniRAD} infrared radiometer,'' {DLR}
  project page, accessed: Sep. 16, 2026.

\bibitem{ESAACTEventVision}
{European Space Agency Advanced Concepts Team}, ``Event-based vision for
  navigation and landing,'' {ESA} Advanced Concepts Team project page, 2023,
  accessed: Sep. 16, 2026.

\bibitem{NASATechPortFlashLidar}
{NASA TechPort}, ``Low {SWaP} flash {LiDAR} for fast traversing rovers,''
  {NASA} TechPort project page, project 113337, accessed: Sep. 16, 2026.

\bibitem{NASASPLICEHDL}
{NASA Science}, ``Hazard detection {Lidar},'' {NASA} Science mission page,
  2024, accessed: Sep. 16, 2026.

\bibitem{Junnosuke2024}
J.~Kamohara \emph{et~al.}, ``Modeling of terrain deformation by a grouser wheel
  for lunar rover simulation,'' arXiv preprint arXiv:2408.13468, 2024.

\bibitem{Jakob2026}
J.~M. Kern \emph{et~al.}, ``Data-driven terramechanics approach towards a
  realistic real-time simulator for lunar rovers,'' in \emph{2025 International
  Conference on Space Robotics (iSpaRo)}, 2025, pp. 662--668.

\bibitem{jpl2022}
{Jet Propulsion Laboratory}, ``{NASA}'s {Mars} helicopter scouts ridgeline for
  perseverance science team,'' May 2022, accessed: Sep. 16, 2026.

\bibitem{Cauligi2025CRESCENT}
A.~Cauligi \emph{et~al.}, ``{CRESCENT}: Collision-free highly constrained
  trajectory optimization for driving on the {Moon},'' \emph{IEEE Transactions
  on Field Robotics}, vol.~3, pp. 177--200, 2026.

\bibitem{fong2022viper}
T.~Fong, ``Volatiles investigating polar exploration rover,'' NASA Technical
  Reports Server (NTRS), Nov. 2022, presentation, November 2, 2022. Document
  ID: 20220016618.

\bibitem{McHenry2020Mars2020AutoNav}
M.~McHenry \emph{et~al.}, ``{Mars} 2020 autonomous rover navigation,'' in
  \emph{43rd Annual AAS Guidance, Navigation and Control Conference}, Jan.
  2020, paper AAS 20-101.

\end{thebibliography}
}

\end{document}